\documentclass[letterpaper, 10 pt, conference]{ieeeconf}  

\IEEEoverridecommandlockouts                              

\usepackage{graphicx}
\usepackage{multirow}
\usepackage{multicol}
\usepackage{comment}
\usepackage{breqn}
\usepackage[top=20.1truemm,bottom=15.2truemm,left=16.9truemm,right=16.9truemm]{geometry}

\title{\LARGE \bf
Spatial Attention Point Network for Deep-learning-based Robust Autonomous Robot Motion Generation
}

\author{ Hideyuki Ichiwara$^{1,\dag}$, Hiroshi Ito$^{1,2,\dag}$, Kenjiro Yamamoto$^{1}$, Hiroki Mori$^{2}$ and Tetsuya Ogata$^{2}$
\thanks{$^{1}$Hideyuki Ichiwara, Hiroshi ito and Kenjiro Yamamoto are with Center for Technology Innovation - Mechanical Engineering, Research \& Development Group, Hitachi, Ltd., Ibaraki, 312-0034, Japan
        {\tt\small hideyuki.ichiwara.bn@hitachi.com}}
\thanks{$^{2}$Hiroshi Ito, Hiroki Mori and Tetsuya Ogata are with Department of Intermedia Art and Science School of Fundamental Science and Engineering, Waseda University, Tokyo, 169-855, Japan
        {\tt\small ogata@waseda.jp}}
\thanks{\dag Both authors contributed equally}
}

\begin{document}

\maketitle
\thispagestyle{empty}
\pagestyle{empty}

\begin{abstract}
Deep learning provides a powerful framework for automated acquisition of complex robotic motions. However, despite a certain degree of generalization, the need for vast amounts of training data depending on the work-object position is an obstacle to industrial applications. Therefore, a robot motion-generation model that can respond to a variety of work-object positions with a small amount of training data is necessary. In this paper, we propose a method robust to changes in object position by automatically extracting spatial attention points in the image for the robot task and generating motions on the basis of their positions. We demonstrate our method with an LBR iiwa 7R1400 robot arm on a picking task and a pick-and-place task at various positions in various situations. In each task, the spatial attention points are obtained for the work objects that are important to the task. Our method is robust to changes in object position. Further, it is robust to changes in background, lighting, and obstacles that are not important to the task because it only focuses on positions that are important to the task.
\end{abstract}

\section{Introduction}
With the increasing automation of lines in the logistics and manufacturing industries, the need for robotic picking actions is increasing. Until now, picking actions in industrial robots have been accomplished by limiting the types and shapes of objects to be picked to a small number of objects or by targeting rigid objects. For example, simple algorithms such as template matching \cite{tsai2003evaluation,barrow1977parametric} are used for industrial bin picking, without the need for advanced object-recognition techniques. To grasp objects of various shapes and types, an advanced object-recognition technique is required.

As an advanced object-recognition technique, deep learning has attracted significant attention. Convolutional neural networks (CNNs) have achieved very high accuracy in the field of image recognition by stacking multiple convolutional and pooling layers for feature extraction and dimensionality compression \cite{krizhevsky2012imagenet,simonyan2013deep,szegedy2015going}. In a robotic application of deep learning, Saxena et al. \cite{lenz2015deep} achieved picking of household items
by estimating the grasping points from RGB images. Levine et al. \cite{levine2018learning} achieved picking of a variety of commodities with a success rate of about 80\% by training up to 800,000 grasps using deep reinforcement learning. Mahler et al. \cite{mahler2017dex} achieved robotic object grasping with 3D CAD simulations to reproduce objects of various shapes and orientations and to predict the grasping positions for thousands of different objects. All of these methods have enabled versatile object grasping, but they have some flaws in terms of accuracy, success rate, and learning cost.

Imitation learning is a method for achieving flexible behavior with low learning costs. Yang et al. \cite{yang2016repeatable} successfully manipulated irregularly shaped objects by remotely teaching a robot to move and training a deep neural network (DNN), which has been difficult to achieve in the past. Ito et al. \cite{ito2020visualization} showed that visualization
by DNNs trained by mimicking multimodal information resulted in the automatic acquisition of features for task execution. However, to achieve highly accurate object grasping, training data are required for each position of the object, and such a large amount of motion training is a challenge.

Therefore, a robot motion-generation model that can respond to a variety of work-object positions with a small amount of training data is necessary. In this paper, we propose a method robust to changes in the object's position by automatically extracting spatial attention points in the image and generating motions on the basis of said points. Further, the obtained spatial attention points are used to predict the camera images simultaneously, which enables us to obtain stable spatial attention points. In this paper, we use a real robot to verify our method’s robustness to changes in the object's position and situation (e.g. background, lighting and obstacles), in a picking task and a pick-and-place task. We also compare our approach to a number of baselines that are representative of previously proposed learning methods.
This paper consists of the following sections. Section \mbox{I\hspace{-.1em}I} describes the challenges CNNs face and related research on robot motion-generation models using CNNs. Section I\hspace{-.1em}I\hspace{-.1em}I describes the proposed motion-generation model, and Section I\hspace{-.1em}V describes the experimental setup. Experimental results and discussion are presented in Section V. We conclude in Section V\hspace{-.1em}I.

\section{RELATED WORK}

\begin{figure*}[t]
  \centering \includegraphics[width=16.0cm]{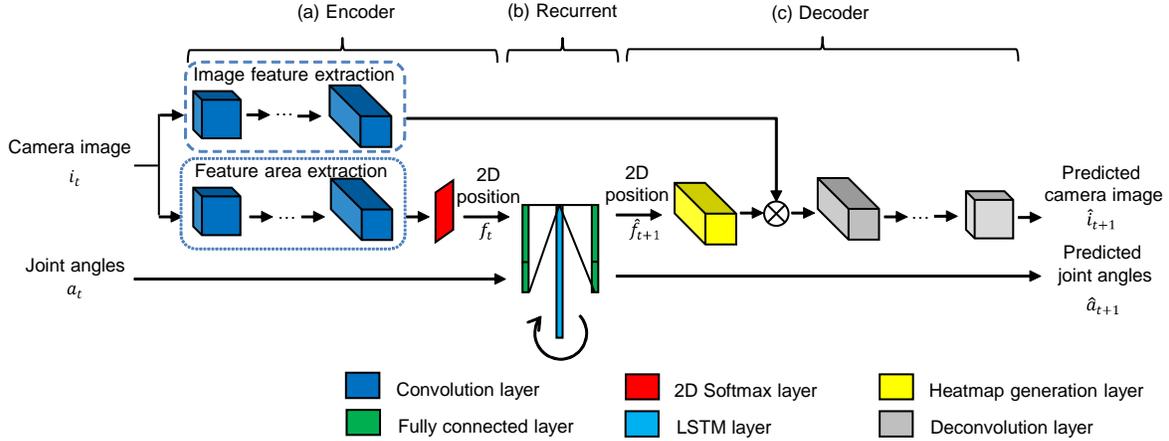}
  \caption{Proposed model architecture (spatial attention point network). The model consists of three parts: an encoder part, a recurrent part, and a decoder part.}
  \label{fig:ModelArchitecture}
\end{figure*}

In general image recognition, highly accurate grasping of an object is achieved by connecting the convolutional layer for feature extraction and the pooling layer for dimensional compression. However, the pooling layer loses information on the spatial structure of an object, although it gains positional invariance \cite{sabour2017dynamic}. Therefore, the collection of vast amounts of image data (e.g., position, angle, rotation, etc.) and data augmentation have been used to achieve robust image recognition. However, end-to-end learning for robots is problematic in that collecting a variety of data requires a large number of trials, which places a heavy burden on the hardware.

In CNNs, CoordConv \cite{DBLP:journals/corr/abs-1807-03247} is a way to account for spatial position information in images. When the image is convolved, a channel representing the x-axis coordinates in the image and a channel representing the y-axis coordinates are added as information about the positional relationship between explicit neurons. The performance is also improved by adding a channel representing the distance from the center of the image. This enables location information to be considered in the image. For example, it is possible to generate an image from two-dimensional coordinates or to extract two-dimensional coordinates from an image. CoordConv is reported to contribute to improved performance in the field of image processing, such as object detection \cite{wang2020solov2}.

For efficient learning of the robot's task, there is a model called a deep spatial autoencoder (DSAE) \cite{finn2016deep} 
that uses positional information in the image to represent the state associated with the robot's task. This model is capable of outputting the positional coordinates of notable feature points in the input image. By multiplying the output values of the CNN by 2D Softmax, the position coordinates of the highest value in each channel are extracted. This enables us to handle explicit location information and extract task-relevant points such as robot arms and objects. However, the extracted position coordinates do not always result in points of attention at the appropriate locations, and a Kalman filter had to be designed because of the noisy nature of the extracted position coordinates.

In this paper, we propose a motion-generation algorithm with location-generalization performance using end-to-end learning. Our method uses 2D Softmax to extract task-critical location information explicitly and learn the time-series relationship between its position coordinates and the robot's joint angles. Further, we use the positional information to predict future camera images to obtain a stable spatial attention point. The details of the motion-generation algorithm are described in the next section.

\section{PROPOSED METHOD}

The structure of the proposed model is shown in Figure \ref{fig:ModelArchitecture}. The model is similar to that proposed by Yang et al. \cite{yang2016repeatable}. It consists of three parts: an encoder part that extracts image features from the camera images, a recurrent part that learns the relationship between the image features and the robot's joint angles, and a decoder part that predicts and reconstructs the camera images at the next time step from the image features and the robot's joint angles. The camera image $i_t$ and robot's joint angle $a_t$ at time $t$ are inputted to the model, and the camera image $\hat{i}_{t+1}$ and robot's joint angle $\hat{a}_{t+1}$ at the next time step $t+1$ are predicted.
\subsection{Encoder part}

In the encoder part, image features and spatial attention points important to the task are output from the input images. The input is a 64$\times$64-pixel RGB image in this study. In the encoder part, there are two CNNs blocks and they do not share the weight. The upper part (Image feature extraction block) shown in the dashed line is for extracting image features and is connected to the decoder part. The lower part (Feature area extraction block) shown in the dotted line is for extracting the feature area of the image and inputs the output to the 2D Softmax function and extracts the positional coordinates of the spatial attention points from the feature map. The number of spatial attention points corresponds to the number of feature maps (number of channels) output from the feature area extraction block.

In addition, the CNN learns the absolute position information of the image covertly, and padding is important for learning the location information \cite{Islam2020HowMP}. Therefore, in our proposed method, we assume that the padding of the CNN is valid.

\subsection{Recurrent part}
In the recurrent part, the relationship between the spatial attention points extracted by the encoder part and the robot's joint angle is learned. The spatial attention points and the robot's joint angles extracted by the encoder section are input to the recurrent part, and the robot's predicted joint angles at the next time step and the spatial attention points for image prediction are output.

\subsection{Decoder part}

In the decoder part, the image is reconstructed from the points output from the recurrent part. The output points are used to generate a heat map centered on the output points. This indicates the region of the image that is important for image prediction. In addition, the feature map generated by the image feature extraction block is multiplied and the weighted feature map is input to deconvolution neural networks (DCNNs) to predict the image at the next time step.
\subsection{Loss function}

The loss function $g$ is shown in Eq. \ref{eq:TotalLoss}. $g_i$ and $g_a$ are the prediction errors of the camera image and the robot's joint angle, expressed in Eq. \ref{eq:ImgLoss} and \ref{eq:AngleLoss}, respectively. $n$ and $m$ are the dimensions of the camera image and the joint angle of the robot, respectively. In addition, $g_f$ is the error of the Euclidean distance between the spatial attention points $f_{t}$ of the encoder output and the spatial attention points $\hat{f}_{t+1}$ of the decoder input, $l$ is the dimensions of the spatial attention points, and $\alpha$ is the weight of the loss function.

%
Spatial attention points of the encoder output are important for task execution, and spatial attention points of the decoder input are important for predicting the camera image. The arm and the object are moved and the camera image changed by task execution, so the spatial attention points of the encoder output and decoder input are expected to be close to each other. By adding $g_f$, stable spatial attention points are obtained.

\begin{eqnarray}
  g&=&g_{i}({i}_{t+1},{\hat{i}}_{t+1}) + g_{a}({a}_{t+1},{\hat{a}}_{t+1}) \nonumber \\
  &&+\alpha g_{f}({f}_{t},{ \hat{f}}_{t+1}) \label{eq:TotalLoss} \\
  g_{i}&=&\frac{1}{n}\sum_{j=1}^{n} ( \hat{i}_j^{t+1} - i_j^{t+1} )^2 \label{eq:ImgLoss} \\
  g_{a}&=&\frac{1}{m}\sum_{k=1}^{m} ( \hat{a}_k^{t+1} - a_k^{t+1} )^2 \label{eq:AngleLoss} \\
  g_{f}&=&\frac{1}{l}\sum_{l=1}^{p} ( \hat{f}_l^{t} - f_l^{t+1} )^2 \label{eq:FeatureLoss} 
\end{eqnarray}

\section{EXPERIMENTS}
\subsection{Hardware}

\begin{figure}[t]
  \centering\includegraphics[width=8.0cm]{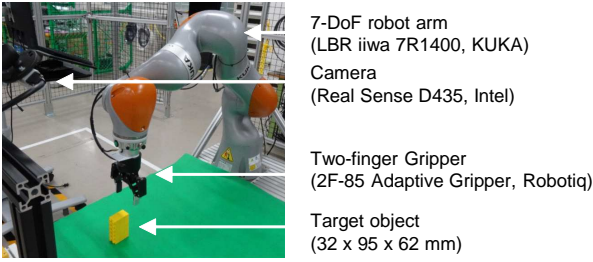}
  \caption{Overview of robot for experiments.}
  \label{fig:RobotOverview}
\end{figure}

\begin{figure}[t]
  \begin{minipage}{4cm}
    \includegraphics[width=4.0cm]{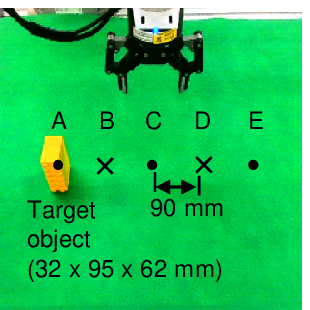}
    \centering\hspace{1.6cm} {\scriptsize (a) Picking task}
    \vspace{0.2cm}
  \end{minipage}
  \begin{minipage}{4cm}
    \includegraphics[width=4.2cm]{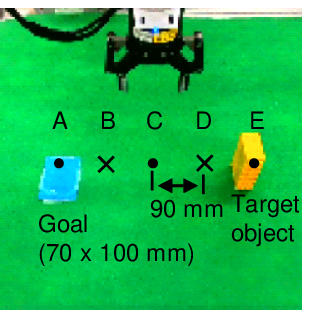}
    \centering\hspace{1.6cm} {\scriptsize (b) Pick and place task}
    \vspace{0.2cm}
  \end{minipage}
  \centering\includegraphics[width=6.5cm]{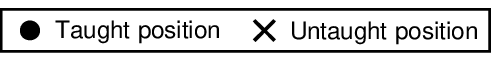}
  \caption{Experimental conditions for picking task and pick-and-place task. (a) is the picking task: to grasp a yellow block. (b) is the pick-and-place task: to grasp a yellow block and place it on a blue area.}
  \label{fig:ExperimentalCondition}
\end{figure}

\begin{figure}[t]
  \begin{minipage}{2cm}
    \includegraphics[width=2.0cm]{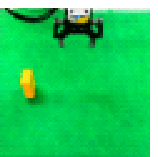}
    \centering\hspace{1.6cm} {\scriptsize $\rm(\hspace{.18em}i\hspace{.18em})$}
  \end{minipage}
  \hfill
  \begin{minipage}{2cm}
    \includegraphics[width=2.0cm]{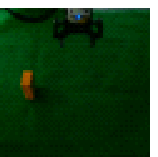}
    \centering\hspace{1.6cm} {\scriptsize $\rm(\hspace{.08em}ii\hspace{.08em})$}
  \end{minipage}
  \begin{minipage}{2cm}
    \includegraphics[width=2.0cm]{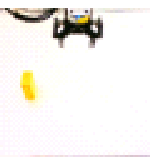}
    \centering\hspace{1.6cm} {\scriptsize $\rm(i\hspace{-.08em}i\hspace{-.08em}i)$}
  \end{minipage}
  \hfill
  \begin{minipage}{2cm}
    \includegraphics[width=2.0cm]{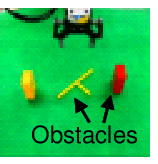}
    \centering\hspace{1.6cm} {\scriptsize $\rm(i\hspace{-.08em}v\hspace{-.06em})$}
  \end{minipage}
  \caption{Experimental situations. $\rm(\hspace{.18em}i\hspace{.18em})$ is the same as training. $\rm(\hspace{.08em}ii\hspace{.08em})$ is with a lighting change. $\rm(i\hspace{-.08em}i\hspace{-.08em}i)$ is with a background change. $\rm(i\hspace{-.08em}v\hspace{-.06em})$ is with an obstacle.}
  \label{fig:ExperimentalConditionTask1}
\end{figure}

Figure \ref{fig:RobotOverview} shows the experimental apparatus used in this paper. The robot arm is KUKA LBR iiwa 7R 1400 and the gripper is a Robotiq 2F-85 Adaptive Gripper. The camera is placed in front of the arm and uses Intel RealSense D435.

When collecting training data, the robot arm was operated by a human using a joystick. The camera image data and robot arm's joint angle data were used as the training data.


We conduct training the models on the ABCI (AI Bridging Cloud Infrastructure) supercomputer operated by the National Institute of Advanced Industrial Science and Technology (AIST) in Japan. Due to the large network size of the proposed model, the training required large computational resources with large GPU memory. The ABCI system is 16 GB/1 node and 2 nodes were used in this study.

\subsection{Task and Dataset}


Experiments were conducted on the following two tasks to evaluate the robustness of the proposed method to changes in object position and situation.

\vspace{\baselineskip}
\subsubsection{Picking task}

Figure \ref{fig:ExperimentalCondition}(a) shows the experimental situation of the picking task. A yellow block (32 x 95 x 62 mm) was used as the grasped object. Five positions were selected at 9 cm intervals, and the motion data at points A, C, and E (designated with black dots) were collected four times each. We evaluated the task success rate after ten trials when blocks were placed at each point, A–E. For comparison with the proposed method, we used 2D Softmax with long short-term memory (LSTM) \cite{DBLP:journals/corr/LevineFDA15}, DSAE with LSTM \cite{finn2016deep} and multimodal recurrent autoencoder (MRAE) \cite{ito2020visualization}. The same experiments were conducted for the three models.

In addition, experiments were conducted with the four situations shown in Figure \ref{fig:ExperimentalConditionTask1} to evaluate the robustness of the proposed method. Figure \ref{fig:ExperimentalConditionTask1}$\rm(\hspace{.18em}i\hspace{.18em})$ shows the same situation as the training, 
$\rm(\hspace{.08em}ii\hspace{.08em})$ shows the situation with darker lighting, $\rm(i\hspace{-.08em}i\hspace{-.08em}i)$ is the situation with a changed background (table color), and $\rm(i\hspace{-.08em}v\hspace{-.06em})$ is the situation with obstacles (blocks of the same shape and different colors than the object and objects of different shapes with the same color).
\vspace{\baselineskip}
\subsubsection{Pick-and-place task}

To verify the applicability to more complex tasks, we conducted experiments with a pick-and-place task. Figure \ref{fig:ExperimentalCondition}(b) shows the experimental situation of the pick-and-place task. The same blocks as in the picking task were used in the pick-and-place task, and the place position was set on a blue sheet (70 $\times$ 100 mm). The blocks and place positions were selected at 5 points at 90 mm intervals. The motion data at points A, C, and E (designated with black dots) were collected three times each for the training data, while points B and D (designated with $\times$ marks) were not taught. We evaluated the success rate of the task with the blocks and place positions at the points A–E after ten trials. Further, to evaluate the robustness of the proposed method, we conducted experiments in the four situations shown in Figure \ref{fig:ExperimentalConditionTask1}.

\section{RESULTS}
\subsection{Picking task}
\subsubsection{Evaluation of robustness to position change}

\begin{table}[t]
\begin{center}
 \caption{Success rate of picking task for each model.}
  \begin{tabular}{|l||c|c|c|} \hline
    Model & Taught & Untaught &All(99\%CI) \\ \hline
    2D Softmax + LSTM\cite{DBLP:journals/corr/LevineFDA15}&66.7\%&25.0\%&48.0$\pm{18.2}$\%\\ \hline
    DSAE + LSTM\cite{finn2016deep}&86.7\%&10.0\%&56.0$\pm{18.1}$\%\\ \hline
    MRAE\cite{ito2020visualization}&96.7\%&10.0\%&62.0$\pm{17.9}$\%\\ \hline
    SPAN (Proposed model)     &100.0\%&100.0\%&100-13.9\% \\ \hline 
  \end{tabular}
  \label{tab:Task1SucessRateMethod}
\end{center}
\end{table}

\begin{figure}[t]
  \centering \includegraphics[width=8.0cm]{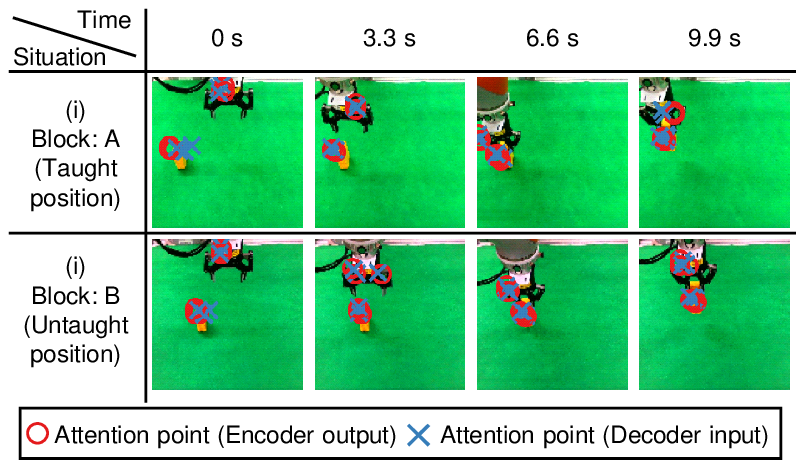}
  \caption{Visualizing attention point with input camera image during execution of picking task.}
  \label{fig:Task1SnapshotsPos}
\end{figure}

\begin{figure}[htbp]
  \begin{minipage}{4cm}
    \centering\includegraphics[width=4.0cm]{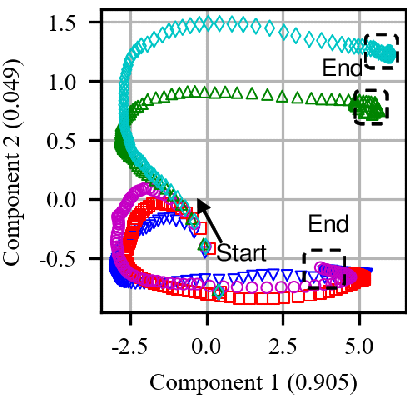}
    \centering\hspace{1.6cm}{\scriptsize (a) 2D Softmax + LSTM}
  \end{minipage}
  \hfill
  \begin{minipage}{4cm}
    \centering\includegraphics[width=4.0cm]{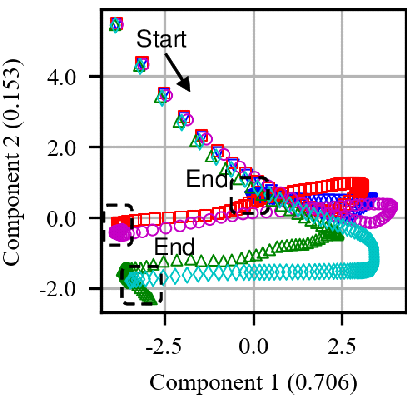}
    \centering\hspace{1.6cm}{\scriptsize (b) DSAE + LSTM}
  \end{minipage}
  \begin{minipage}{4cm}
    \vspace{0.5cm}
    \centering\includegraphics[width=4.0cm]{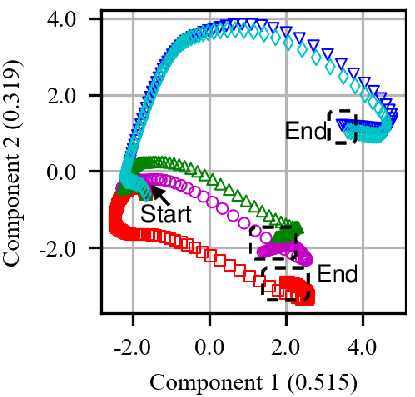}
    \centering\hspace{1.6cm} {\scriptsize (c) MRAE}
  \end{minipage}
  \hfill
  \begin{minipage}{4cm}
    \vspace{0.5cm}
    \centering\includegraphics[width=4.0cm]{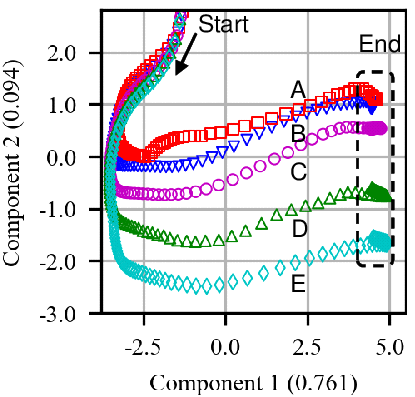}
    \centering\hspace{1.6cm} {\scriptsize (d) Proposed model (SPAN)}
  \end{minipage}
  
  \centering\vspace{0.5cm}\includegraphics[width=6.5cm]{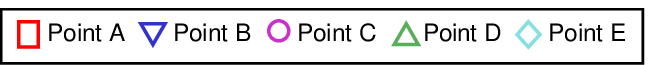}
  \caption{Internal state of LSTM in models during execution of picking task.}
  \label{fig:Task1Attracter}
\end{figure}

Table \ref{tab:Task1SucessRateMethod} shows the task success rates for the taught and untaught positions using each model. For the taught positions, all methods have a success rate of more than 60\%, but for the untaught positions, the success rate was less than 25\% except for the proposed model. The proposed model showed a 100\% success rate for both the taught and untaught positions, and the proposed model was robust to changes in the object position.

Figure \ref{fig:Task1SnapshotsPos} shows an example of the arm motion and spatial attention points for the proposed model. The images are shown at 3.3-s intervals, i.e. the time elapsed between each image from left to right is 3.3 s. The red circles are the spatial attention points output from the encoder part, and the blue x marks are the spatial attention points output from the recurrent part. The spatial attention points are shown for teaching position A and untaught position B. There are attention points in the robot hand and the target object. These points are important for the grasping task and are extracted appropriately. Note that spatial attention points are on the object even if the object is in an untaught position.

Next, the internal state of the LSTM was visualized by principal component analysis (PCA) to verify the effectiveness of the models with respect to the change in position of the object. Figure \ref{fig:Task1Attracter} shows the internal state of the LSTM for each model. The principal components of the recursive coupling layer of the LSTM are represented as two dimensionless axes, and each point is the internal state of the LSTM at each operation (100 plots for 10 s every 100 ms from the start of the operation). For the models in (a), (b) and (c), the location of the object is different, but the internal state of the LSTM is often the same. For example, for the MRAE in (c), the internal state is switched for the taught positions A, C, and E, but it is the same for the untaught position B and the position E. For the MRAE in (c), the internal state of the LSTM is the same for the untaught position B and the taught position E. The actual arm motion was also the same when the object was at point B and at point E, and the grasping failed. However, for the proposed model in (d), the internal state of the LSTM is switched in response to the position of the block. For example, the untaught point B is located between taught points A and C. The internal state is also intermediate between taught points A and C in the same way. These results suggest that the proposed model is able to become robust to position changes by switching the internal state of the LSTM in accordance with the object position.

\subsubsection{Evaluation of robustness to Situational change}


\begin{table}[t]
\begin{center}
 \caption{Success rate of picking task for proposed model in each situation.}
  \begin{tabular}{|l||c|c|c|} \hline
    Situation & Taught pos. & Untaught pos. &All(99\%CI) \\ \hline
    $\rm(\hspace{.18em}i\hspace{.18em})$        &100.0\%&100.0\%&100.0-13.9\%\\ \hline
    $\rm(\hspace{.08em}ii\hspace{.08em})$        &100.0\%&100.0\%&100.0-13.9\%\\ \hline
    $\rm(i\hspace{-.08em}i\hspace{-.08em}i)$ &100.0\%&100.0\%&100.0-13.9\%\\ \hline
    $\rm(i\hspace{-.08em}v\hspace{-.06em})$                    &100.0\%&100.0\%&100.0-13.9\%\\ \hline
  \end{tabular}
  \label{tab:Task1SucessRateEnv}
\end{center}
\end{table}

\begin{figure}[t]
  \centering \includegraphics[width=8.0cm]{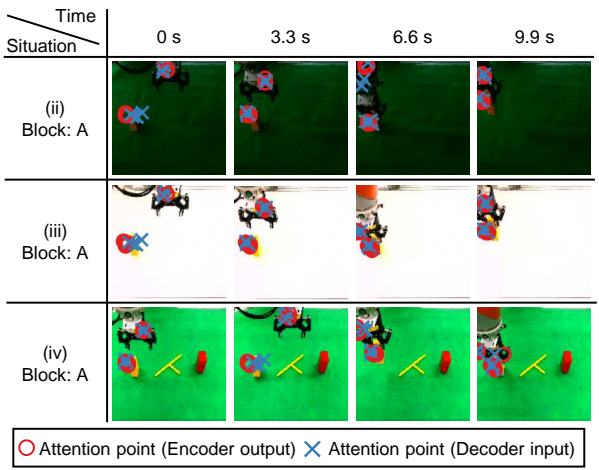}
  \caption{Visualizing attention point with input camera image during execution of picking task in each situation.}
  \label{fig:Task1SnapshotsRobust}
\end{figure}

Table \ref{tab:Task1SucessRateEnv} shows the task success rates for taught and untaught positions in each case of the proposed model. In all cases, the success rate was 100\% and did not change regardless of the situation. Figure \ref{fig:Task1SnapshotsRobust} shows an example of arm movements and spatial attention points at each situation. From top to bottom, situation $\rm(\hspace{.08em}ii\hspace{.08em})$, when the block position is at taught point A, as well as situation $\rm(i\hspace{-.08em}i\hspace{-.08em}i)$ and $\rm(i\hspace{-.08em}v\hspace{-.06em})$, are shown. In each case, as in situation $\rm(\hspace{.18em}i\hspace{.18em})$, we show that spatial attention points are in the vicinity of the hand and the object. Less important to task execution, the hand is not paying attention to the background or obstacles and is considered robust to those changes. 

\subsection{Pick-and-place task}

\begin{table}[t]
\begin{center}
 \caption{Success rate of pick-and-place task for proposed model in each situation.}
  \begin{tabular}{|l||c|c|c|c|c|} \hline
    &Taught&Taught&Untaught&Untaught& \\
    Situ.&to&to&to&to&All(99\%CI) \\
    &taught&untaught&taught&untaught& \\ \hline
    $\rm(\hspace{.18em}i\hspace{.18em})$&85.0\%&81.7\%&100.0\%&85.0\%&$88.5\pm{5.8}\%$ \\ \hline
    $\rm(\hspace{.08em}ii\hspace{.08em})$&95.0\%&85.0\%&100.0\%&95.0\%&$93.5\pm{4.5}\%$ \\ \hline
    $\rm(i\hspace{-.08em}i\hspace{-.08em}i)$&83.7\%&83.7\%&100.0\%&80.0\%&$88.0\pm{5.9}\%$ \\ \hline
    $\rm(i\hspace{-.08em}v\hspace{-.06em})$&85.0\%&78.4\%&100.0\%&90.0\%&$88.0\pm{5.9}\%$ \\ \hline
  \end{tabular}
  \label{tab:Task2SucessRate}
\end{center}
\end{table}
\begin{figure}[t]
  \centering \includegraphics[width=8.0cm]{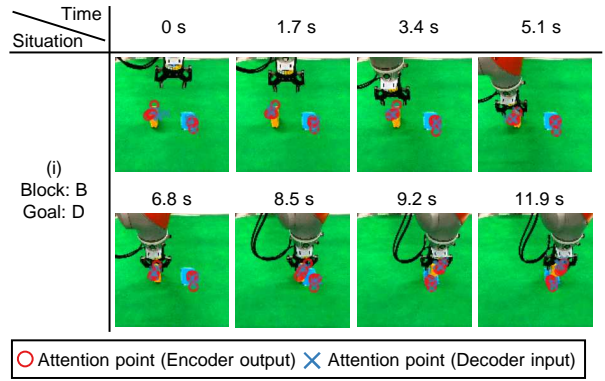}
  \caption{Visualizing attention point with input camera image during execution of pick-and-place task.}
  \label{fig:Task2Snapshots}
\end{figure}
\begin{figure}[t]
  \begin{minipage}{4cm}
    \includegraphics[width=4.0cm]{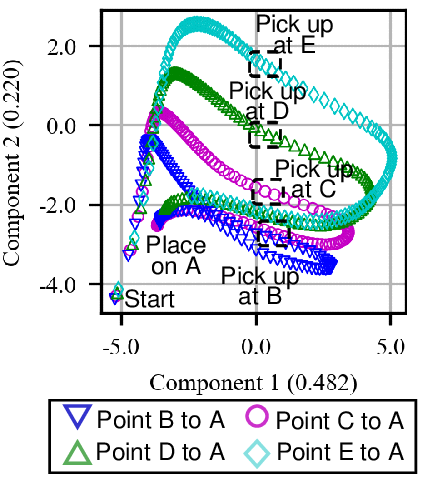}
    \centering\hspace{1.6cm} {\scriptsize (a) Place on A}
  \end{minipage}
  \begin{minipage}{4cm}
    \includegraphics[width=4.0cm]{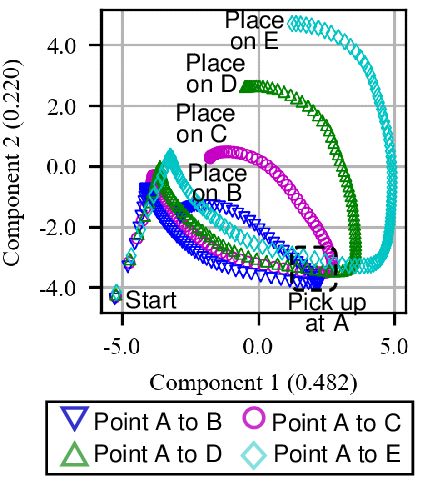}
    \centering\hspace{1.6cm} {\scriptsize (b) Pick up at A}
  \end{minipage}
  \caption{Internal state of LSTM in proposed model during execution of pick-and-place task.}
  \label{fig:Task2Attracter}
\end{figure}

Table \ref{tab:Task2SucessRate} shows the task success rate in each case for the proposed model. We categorize the cases into taught position to taught position, taught position to untaught position, untaught position to taught position, and untaught position to untaught position. The task success rates are equally high in the cases with and without the inclusion of an untaught position.

Figure \ref{fig:Task2Snapshots} shows an example of the arm motion and spatial attention points in the proposed model. The block is at the untaught position B, and the goal is at the untaught position D. The images are shown at 1.7-s intervals, i.e. the time elapsed between each image from left to right is 1.7 s. There are spatial attention points in the object, the block, and the goal, and we were able to extract the change in the object's position.

Figure \ref{fig:Task2Attracter} shows the internal state of the LSTM in the proposed model when the blocks are picked and placed (a) from positions B–D to A and (b) from position A to B–D. In (a), the internal state is switched from the initial state in accordance with the block position, and the internal state converges to a single point because the goal position is the same. In (b), the internal state is the same until the arm is moved to the same block position A, and the internal state is switched in accordance with the goal position. These results suggest that the proposed model is able to switch the internal state of the LSTM in accordance with the position of the object and achieve a motion that is robust to changes in position, even in tasks more complex than the picking task.



In addition, the overall task success rate is robust to situational changes, with the overall task success rate remaining the same (around 90\%) regardless of situation.

\section{CONCLUSION}

In this paper, we proposed a method robust to changes in the position of objects by automatically extracting spatial attention points, such as the position of the object or the robot hand in the image that is important for the task, and generating motions on the basis of the spatial attention points. To obtain stable spatial attention points, we also proposed to simultaneously predict camera images with using the obtained spatial attention points. To verify the effectiveness of the proposed model, experiments were conducted using a real robot for a picking task and a pick-and-place task. In these experiments, we were able to obtain spatial attention points for the objects and hands and confirmed that the model is robust to positional changes. In addition, we confirmed that the system is robust to changes in the background, lighting, and obstacles in a situation where there is no deterioration in performance. This is because the spatial attention points are only in the areas that are important for the task and not in the backgrounds that are unimportant for the task. In conclusion, we confirmed that the proposed model is robust to changes in object position and situation.

A task for future work is to train the recurrent part on the basis of the image features so that we can consider not only the position of the object but also the orientation of the object and the arm. This work is necessary because complex tasks may be difficult to perform on the basis of only the position of the object in the image.

\section*{Acknowledgement}
Computational resource of AI Bridging Cloud Infrastructure (ABCI)
provided by National Institute of Advanced Industrial Science and
Technology (AIST) was used.

\bibliographystyle{IEEEtran}
\bibliography{ref}

\end{document}